\pdfoutput=1
 \pdfoutput=1
 \pdfoutput=1
 \pdfoutput=1
 \pdfoutput=1
\documentclass[a4paper,fleqn]{cas-dc}
\usepackage[numbers]{natbib}
\usepackage{amsmath}

\newcommand{\myref}[1]{%
  \hyperref[#1]{Eq.~(\ref*{#1})}%
}
\makeatletter
\renewcommand{\fnum@figure}{Fig. \thefigure.\@gobble}
\makeatother

\usepackage{cleveref}

\usepackage{hyperref}
\hypersetup{hypertex=true,
            colorlinks=true,
            linkcolor=blue,
            anchorcolor=blue,
            citecolor=blue}
\usepackage{graphicx}
\usepackage{CJKutf8}
\usepackage{subfigure}
\usepackage{epstopdf}
\usepackage{xcolor}
\usepackage{cleveref}
\usepackage{caption}
\usepackage{float} 
\usepackage{utfsym}
\usepackage{pifont}
\usepackage[noend]{algpseudocode}
\usepackage{algorithmicx,algorithm}

\def\tsc#1{\csdef{#1}{\textsc{\lowercase{#1}}\xspace}}
\tsc{WGM}
\tsc{QE}
\tsc{EP}
\tsc{PMS}
\tsc{BEC}
\tsc{DE}

\begin{document}
\begin{CJK}{UTF8}{gbsn}
\begin{sloppypar}
	\let\WriteBookmarks\relax
	\def\floatpagepagefraction{1}
	\def\textpagefraction{.001}
	\let\printorcid\relax
	\shorttitle{}
	\shortauthors{S. Fan et~al.}

	\title [mode = title]{Addressing Domain Discrepancy: A Dual-branch Collaborative Model to Unsupervised Dehazing}
	\author[1]{\textcolor[RGB]{0,0,1}{Shuaibin Fan}}
	\ead{Fansb@stu.cqut.edu.cn}
	\author[1]{\textcolor[RGB]{0,0,1}{Minglong Xue}}
	\address[1]{College of Computer Science and Engineering, Chongqing University of Technology, Chongqing 400054, China}
 	\address[2]{Industrial Training Center, Guangdong Polytechnic Normal University, Guangzhou, 510665, China}
	\cormark[1]
	\ead{xueml@cqut.edu.cn}
	\author[1]{\textcolor[RGB]{0,0,1}{Aoxiang Ning}}
	\ead{ningax@stu.cqut.edu.cn}
 	\author[2]{\textcolor[RGB]{0,0,1}{Senming Zhong}}
	\ead{itcsmzhong@gpnu.edu.cn}
	\cortext[cor1]{Corresponding author.} 

	\begin{abstract}
Although synthetic data can alleviate acquisition challenges in image dehazing tasks, it also introduces the problem of domain bias when dealing with small-scale data. This paper proposes a novel dual-branch collaborative unpaired dehazing model (DCM-dehaze) to address this issue. The proposed method consists of two collaborative branches: dehazing and contour constraints. Specifically, we design a dual depthwise separable convolutional module (DDSCM) to enhance the information expressiveness of deeper features and the correlation to shallow features. In addition, we construct a bidirectional contour function to optimize the edge features of the image to enhance the clarity and fidelity of the image details. Furthermore, we present feature enhancers via a residual dense architecture to eliminate redundant features of the dehazing process and further alleviate the domain deviation problem. Extensive experiments on benchmark datasets show that our method reaches the state-of-the-art. This project code will be available at \url{https://github.com/Fan-pixel/DCM-dehaze.}
 
	\end{abstract}
	\maketitle
	\section{Introduction}

	\par{
Images with complex, hazy weather environments are often visually compromised with poor visibility~\cite{guo2023haze}. These degraded images usually affect many vision tasks (such as autonomous driving~\cite{sun2023scale,mahaur2023small}, target detection~\cite{ye2024low,li2020netnet,shi2023cnn}). Currently, the main deep learning tasks are based on the processing of clear images. Therefore, the study of dehazing is very important to improve the clarity of the image. Currently, the main single image dehazing methods are divided into two categories: the early dehazing method and the deep learning-based method.

In early studies, most researchers used prior information to dehaze images. Such as dark channel a priori methods~\cite{ref1}, colour attenuation a priori methods~\cite{ref2}, and non-local colour a priori methods~\cite{ref3}. The atmospheric scattering model equations can be written as~\myref{eq11}:
\begin{flalign}\label{eq11}
&\boldsymbol I(x) = \boldsymbol J(x)t(x) + \boldsymbol A(1-t(x)),&
\end{flalign}
where is $ I(x) $ the hazy image, $ J(x) $ the clear image, $ t(x) $ the transmittance, and $ A $ the global atmospheric light. Usually defining $t(x)=\boldsymbol e^{-\beta d(x)}$ .

The $ \beta $ represents the atmospheric scattering coefficient, and $ d(x) $ is the scene depth. \par As can be seen from~\myref{eq11}, image dehazing methods based on atmospheric scattering models require the estimation of $t(x)$ and $A(x)$. However, while improving the visibility of grayscale images, these methods also rely on specific a priori assumptions that are not always practical when dealing with grayscale in real-world images, given the complexity of the real world and its ever-changing nature. \par
\begin{figure}[t]
\centering
\includegraphics[width=0.8\linewidth]{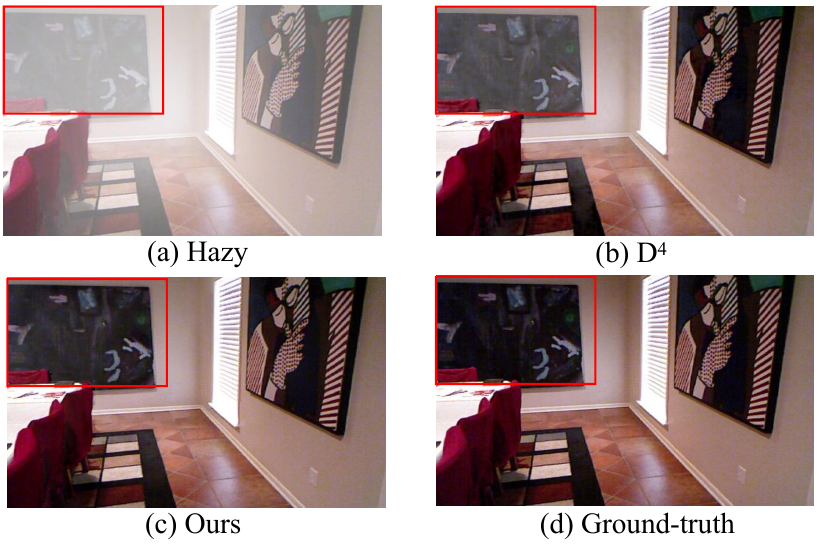}
\caption{A single image haze removal example. (a) The hazy image. (b) Result of Yang et al. \cite{yang2022self} proposed network. (c) The dehazed image of our method. (d) The Ground-truth image.}
\label{fig:fig1}
\end{figure}
Existing studies have demonstrated that direct recovery of image resolution is challenging without the guidance of atmospheric scattering models\cite{guo2022image,wu2021contrastive}. To solve this problem, researchers have proposed numerous methods~\cite{liu2023nighthazeformer,yang2022self} based on deep learning. The $D^4+$~\cite{yang2023robust} focus is exploring the scattering coefficient and depth information in hazy and clean images. However, as shown in ~\autoref{fig:fig1} (b). There is a serious problem of domain deviation in the conversion process. \par
This paper proposes a novel dehaze network (DCM-dehaze) to solve the above problems. The DCM-dehaze framework benefits from adversarial training efforts and generalizes well to real-world hazy scenarios. To better recover clean images from hazy images, we propose a new Dense Flow Residual Enhancer (DFRE) to encourage the removal of redundant features (e.g., rain, haze, noise, etc.) during image restoration. This effectively improves the network's haze-removal effect. In addition, we further explore the problem of detailed feature distortion due to domain bias during image reconstruction. We propose a new depth-separable convolution method and a bidirectional contour analysis technique to solve the domain bias and detail distortion problems in image reconstruction and enhance the image details. The main contributions of this work are summarized as follows:

\renewcommand{\labelitemi}{\scalebox{0.8}{$\bullet$}} 
\begin{itemize}
  \setlength{\itemsep}{0pt} 
  \setlength{\parsep}{0pt}

\item We present a new unsupervised dehazing network of the proposed DCM-dehaze executed without pairwise training information, which can effectively address the recovery of invisible, real-world hazy scenes.
\item To effectively eliminate redundant features (e.g., rain, haze, noise, etc.) and to obtain deeper features, we propose the DFRE and DDSCM modules.
\item Since haze images contain less contrast in the feature fusion phase, we propose a bidirectional contour analysis (BCA) module to enhance edges and textures. It can effectively repair invisible hazy scenes.
\item The superiority of our designed method is demonstrated through extensive experiments on benchmark datasets.
\end{itemize}
	}

	\section{Related work}
	\label{Related_work}
	\par{
		Image dehazing, which aims to generate haze-free images from hazy images, can be broadly classified into prior assumptions and learning-based methods.
	}
	\subsection{Prior-based Image Dehazing Methods}
	\par{
		Early dehazing methods usually explore hand-crafted priors, which mainly utilize information from the image for dehazing. Among them, He~\cite{ref1} proposed Dark Channel Prior (DCP) is a classical approach that uses the dark channel prior in natural scene images to estimate the hazy concentration in the scene, and the non-local prior proposed by Berman~\cite{ref3} for a network of individual image dehazing. However, a new algorithm proposed by Tarel~\cite{tarel2009fast} can solve the haze and ambiguity between objects with low color saturation. In addition, Retinex theory, histogram equalization, and edge-preserving filtering are widely used for single-image dehazing. Although these priori-based methods have achieved good results, they are not applicable in today's complex real-world environment.
	}
	\subsection{Learning-based Image Dehazing Methods}
	\par{
		Learning-based image dehazing methods~\cite{cong2024semi,wang2024ucl,zhang2024depth,hodges2019single,li2019generative} differ from a priori-based methods. The former is data-driven and mainly consists of deep neural networks to estimate the transmittance map and atmospheric light in a physical scattering model~\cite{ren2016single,zhang2018densely}. Work shown by~\cite{li2017aod} proposed an end-to-end AODNet dehazing network that produces clear images by reformulating the atmospheric scattering model. Song et al.~\cite{song2023vision} pointed out the limitations of Swin Transformer in image denoising and proposed DehazeFormer with better performance and lower cost. At present, although there are some methods~\cite{an2022semi,yang2023adversarial,xue2024low} to alleviate the problem of domain deviation to a certain extent, their haze removal ability still needs to be improved, and they over-rely on the quantity and quality of synthetic data.\par
  
  We propose an unsupervised dehazing network designed to address the problem of domain deviation during explicit image reconstruction. We argue that hazy images can also provide useful guidance information. Therefore, our approach utilizes a bidirectional contour optimization algorithm to benefit from real-world images from different domains. Effectively improve the haze removal effect due to excessive reliance on the quantity and quality of synthetic data.
	}
\begin{figure}[htbp]
\centering 
\includegraphics[width=0.98\linewidth]{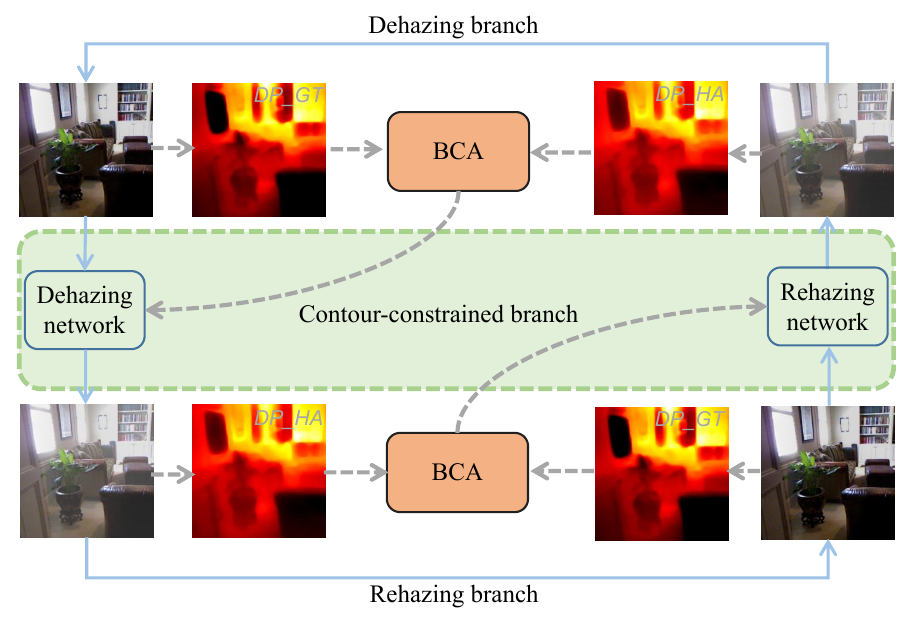}
\caption{The concept of dual-task coordination and mutual promotion. DP-GT and DP-HA denote depth maps of ground-truth and haze removal images. BCA denotes the Bidirectional Contour Analysis module.} 
\label{fig:fig2}
\end{figure}
\section{Methodology}
	\label{Methodology}
In this section, we propose a novel dual-branch collaboration model (DCM-dehaze). Specifically, the model mainly includes the dehaze branch and the contour-constrained branch. The overall framework is shown in~\autoref{fig:fig2}.

\begin{figure*}
\centering 
\includegraphics[width=0.85\textwidth]{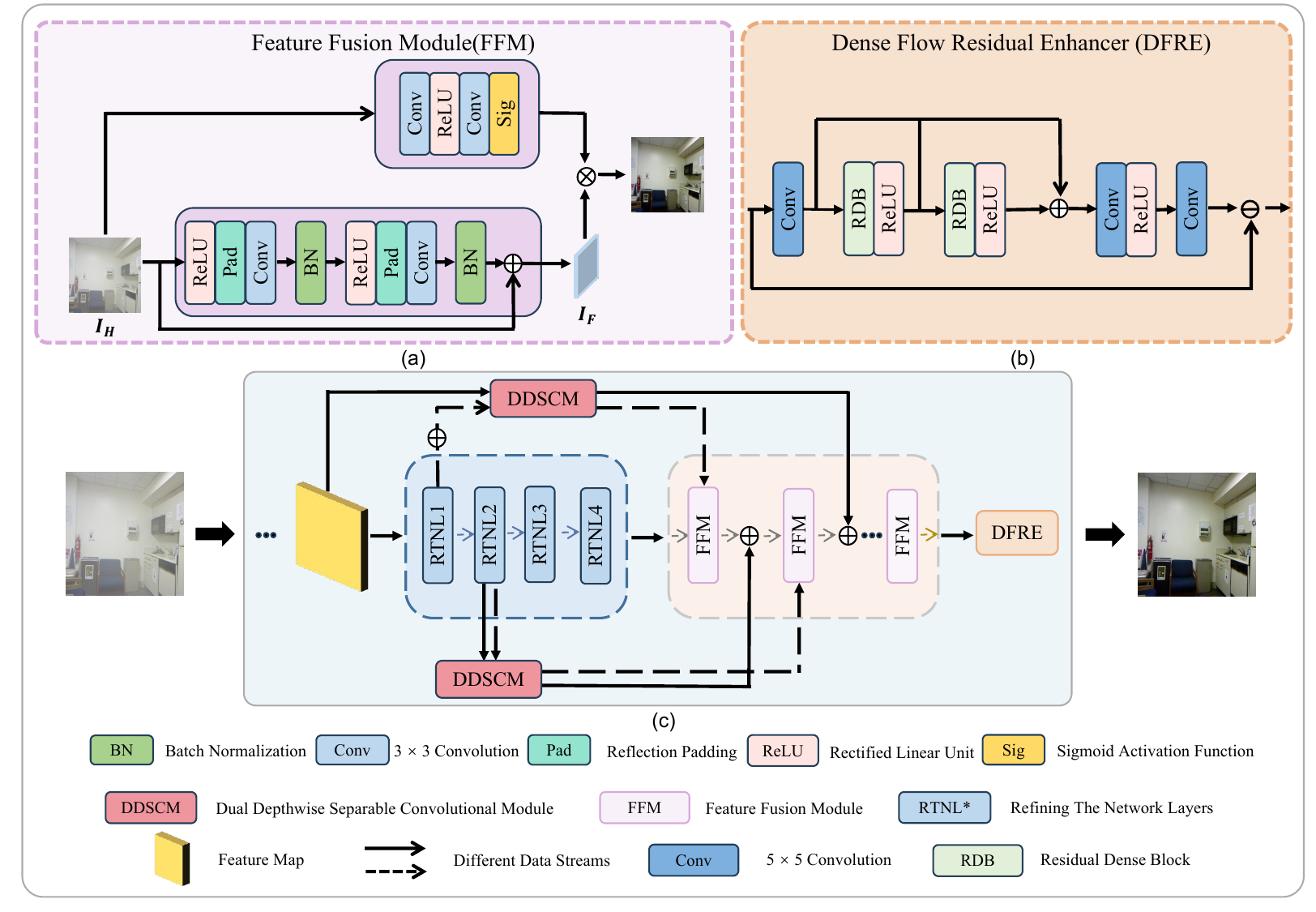} 
\caption{The architecture of the method consists of the dehazing network, DFRE and DDSCM. The FFM is the central fusion part of the dual-task interaction mechanism based on the attention mechanism. This mechanism allows seamless dehazing and contour constraints integration into a unified framework and improves the model's performance by mutual reinforcement.} 
\label{fig:fig3} 
\end{figure*}
\subsection{Dense Flow Residual Enhancer}\label{AA}
Image recovery tasks mostly have known redundant features compared to image enhancement tasks.  For the feature information in the reconstruction process, we designed the DFRE module using the residual-density architecture to remove redundant features with high-frequency information. As is shown in~\autoref{fig:fig3} (b), by combining residual learning and dense connections, DFRE can capture and fuse local and global features of images more efficiently. This design helps improve the model's understanding of complex image structures and enhances the network's ability to learn more subtle feature changes through residual connectivity.\par
\begin{figure}[htbp]
\centering 
\includegraphics[width=0.85\linewidth]{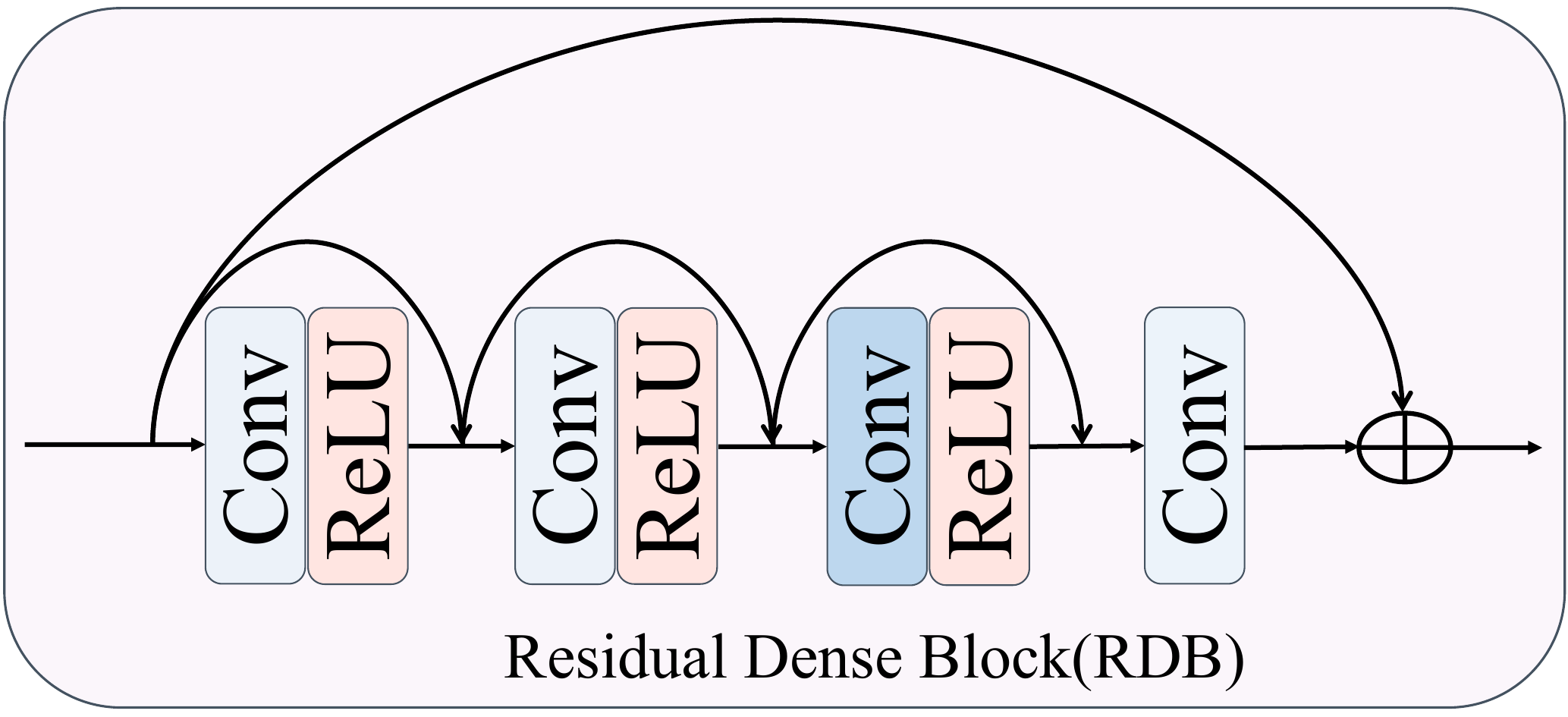}
\caption{The structure of single Residual Dense Block.}
\label{fig:fig4}
\end{figure}
As shown in~\autoref{fig:fig4}, within a residual dense block (RDB), image feature extraction is first enhanced by convolutional layers and residual learning. Then the feature representation is gradually enriched by dense layers and local feature fusion techniques, improving the model's ability to capture details. After two RDB processes and convolutional layer refinement, the features are finally mapped back to the original space and the original inputs are subtracted to highlight key differences.
\subsection{Bidirectional Contour Analysis}
In dehazing networks, contour analysis helps to enhance the edge information of the image, as the dehaze task often results in the image losing some of its detail and clarity, especially for objects in the distance. By optimizing contours, the network can pay more attention to the edge information of the image, thus generating a more transparent and more realistic image. Such an operation helps improve the dehazing network's performance and the quality of the generated images. \par
Specifically, this method is mainly used to obtain the edge information of the image more accurately by calculating the contour of the image along the horizontal and vertical directions. Firstly, the colour image is converted to a grayscale image, and then the horizontal and vertical gradients are calculated for the grayscale image. Then, the magnitude of the gradient in horizontal and vertical directions is calculated, and the mean value of the gradient magnitude is used as the contour loss. This design can effectively guide the model in learning more contour information.
\subsection{Dual Depthwise Separable Convolutional Module}
\begin{figure}[htbp]
\centering
\includegraphics[width=0.36\linewidth]{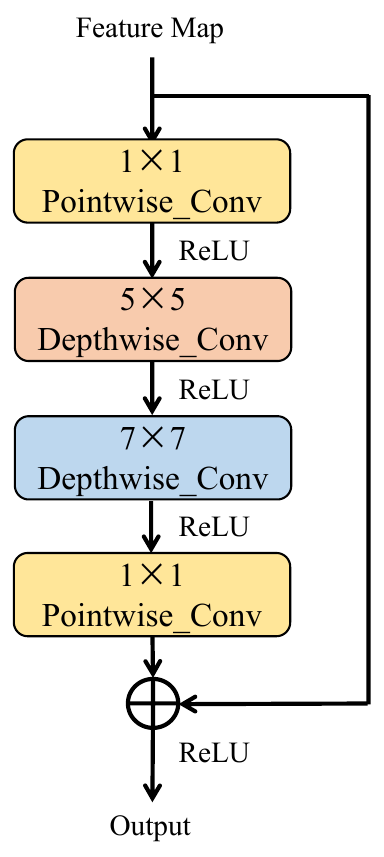}
\caption{The structure of Dual Depthwise Separable Convolutional Module.}
\label{fig:fig5}
\end{figure}

We design a network structure with Dep residual blocks to alleviate the domain bias problem during image reconstruction. As is shown in~\autoref{fig:fig3}~(c), the structure utilizes a bidirectional depth-separable convolutional network, which enhances the network's ability to perceive deep feature information, thus significantly improving the visual effect. This idea comes from ADE-CycleGAN~\cite{yan2023ade} but differs from it to extract deeper information about distant background features and simultaneously capture multi-scale background information. The specific framework is shown in~\autoref{fig:fig5}, meanwhile, It can also be represented by~\myref{eq12}:
\begin{flalign}\label{eq12}
&\boldsymbol I_{temp} =\boldsymbol f_{PW}^{1\times1}\boldsymbol (f_{DW}^{7\times7}\boldsymbol (f_{DW}^{5\times5}\boldsymbol ( f_{PW}^{1\times1}\boldsymbol (I_{input})))))\times I_{input},&
\end{flalign}
where $ f_{PW}^{1 \times 1}  $ is a point-by-point convolution with a convolution kernel size of 1×1, and $ f_{DW}^{5 \times 5} $ and $ f_{DW}^{7 \times 7} $ are channel-by-channel convolutions with convolution kernel sizes of $5 \times 5$ and $7 \times 7$.
\begin{figure*}
\centering 
\includegraphics[width=0.8\textwidth]{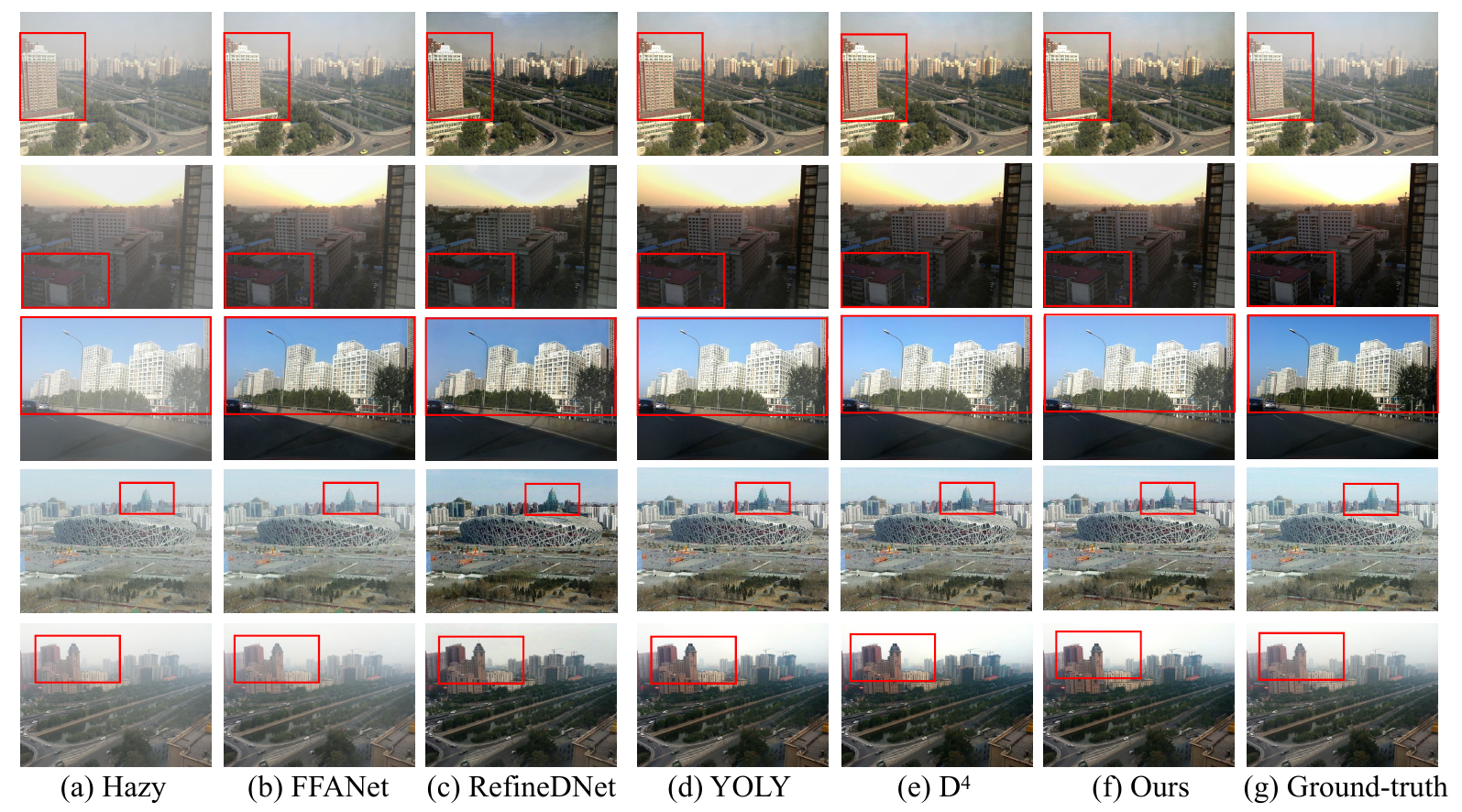} 
\caption{Visual comparison of haze removal on samples from the SOTS-outdoor datasets. All results of FFANet, $D^4$ and YOLY  are hazy. The results of RefineDNet are over-dehazed. Our method dehazes well in all cases.} 
\label{fig:fig6} 
\end{figure*}
\par{As is shown in~\autoref{fig:fig3}~(a), this approach can both effectively fuse image feature information and facilitate the gradual transfer of information to the deeper layers of the network, thereby significantly enhancing the fine feature details of the background image.}
\subsection{Loss Function}
Establishing an effective loss function to bind the dehazing network is important.\par
\textbf{Cycle-consistency loss: } In our dehazing network, let the model reconstructed clear image $\stackrel{\wedge}{G}$ and hazed image $\stackrel{\wedge}{H}$ should be consistent with their input counterparts $G$ and $H$, respectively. The cyclic consistency loss in our dehazing network can be written as~\myref{eq13}:
\begin{flalign}\label{eq13}
&\boldsymbol L_{\mathrm{cyc}}=\boldsymbol E_{G\sim\chi_G}\left\|G-\overset{\wedge}{G}\right\|_1 + \boldsymbol E_{H\sim\chi_H}\left\|H-\overset{\wedge}{H}\right\|_1,&
\end{flalign} 
where $ \left\|\right\|_1 $ is the $L_1$ norm.\par
\textbf{Adversarial learning loss:} To make our dehazing and rehazing images visually realistic and follow the same distribution as the images in the training set $ \chi_H $ and $\chi_G$. The adversarial loss in our dehazing network can be written as:
\begin{flalign}\label{eq1}
&\boldsymbol L_{ad\nu}(\boldsymbol Dis_g)=E[(\boldsymbol Dis_g(\overset{\wedge}{\operatorname*{g}})-1)^2] + \boldsymbol E[ (\boldsymbol Dis_g(\overset{\wedge}{\operatorname*{g}}))^2],&
\end{flalign}
\begin{flalign}\label{eq1}
&\boldsymbol L_{ad\nu}(\boldsymbol G_{dehaze})=E[(\boldsymbol Dis_{g}(\stackrel{\wedge}{g})-1)^2],&
\end{flalign} 
where $ g $ is a true clean sample from the clear image set $ \chi_G $, $ \overset{\wedge}{\operatorname*{g}}$ is the dehazing result from $G_{dehaze}$, and $Dis_g$ is the judgment of whether the input image belongs to the clear image or not.\par
\textbf{Contour Loss:} To encourage the generator to produce images with sharper contours, a bidirectional contours analysis is introduced. The aim is to minimize the blurring of the boundary of objects in the image, thus improving the definition of the edge of the image.
\begin{table*}[width=\textwidth,ht!]
\caption{Quantitative comparisons (Average PSNR/SSIM/CIEDE) with dehazing approaches on the SOTS and I-haze datasets, where AODNet~\cite{li2017aod}, EPDN~\cite{qu2019enhanced}, FFANet~\cite{qin2020ffa}, HardGAN~\cite{deng2020hardgan}, PSD~\cite{chen2021psd} are supervised methods, and CycleGAN~\cite{zhu2017unpaired}, RefinedNet~\cite{zhao2021refinednet}, YOLY~\cite{li2021you},$D^4$~\cite{yang2022self}, $D^4+$~\cite{yang2023robust} are unsupervised methods. The first and second performances are shown in {\color{red} red} and {\color{blue} blue}, respectively.}
\scalebox{0.820}{
\renewcommand\arraystretch{1.2}
\begin{tabular}{llllllllllllll}
\hline
                            & Methods    & Type      & \multicolumn{3}{l}{SOTS-indoor} & \multicolumn{1}{c}{} & \multicolumn{3}{l}{SOTS-outdoor} & \multicolumn{1}{c}{} & \multicolumn{3}{l}{I-haze} \\ \cline{4-6} \cline{8-10} \cline{12-14} 
                            &            &           & PSNR$\uparrow$     & SSIM$\uparrow$     & CIEDE$\downarrow$     &                      & PSNR$\uparrow$      & SSIM$\uparrow$     & CIEDE$\downarrow$     &                      & PSNR$\uparrow$    & SSIM$\uparrow$   & CIEDE$\downarrow$   \\ \hline
\multirow{5}{*}{\centering Paired}     & AODNet~\cite{li2017aod}       & ICCV’2017 & 19.35    & 0.857    & 7.99     &                      & 20.05     & 0.893    & 7.87     &                      & 14.52   & 0.726  & \color{blue}16.01   \\
 & EPDN~\cite{qu2019enhanced}      & CVPR’2019 & 25.06    & 0.915    & 9.960     &             & 20.30     & 0.869   & 10.60     &                      & \color{blue}15.01   & \color{blue}0.744  & 23.76   \\
                            & FFANet~\cite{qin2020ffa}      & AAAI’2020 & \color{red}36.36    & \color{red}0.988    & \color{red}5.77     &                      & \color{blue}24.71     & \color{blue}0.899    & \color{blue}5.27     &                      & 12.01   & 0.589  & 22.57   \\
                             & HardGAN~\cite{deng2020hardgan}      & ECCV’2020 & \color{blue}35.45    & \color{blue}0.986    & \color{blue}5.83     &                      & \color{red}27.84     & \color{red}0.941    & \color{red}3.78     &                      & 13.84   & 0.715  & 19.76   \\
                            & PSD~\cite{chen2021psd}        & CVPR’2021 & 15.02    & 0.764    & 13.80    &                      & 15.63     & 0.834    & 12.60    &                      & \color{red}{15.30}   & \color{red}{0.800}  & \color{red}{14.84}   \\ \hline
\multirow{6}{*}{\centering W/o Paired} & CycleGAN~\cite{zhu2017unpaired}   & ICCV’2017 & 21.33    & 0.880    & 14.70     &                      & 20.55     & 0.845    & 10.29     &                      & 15.70   & \color{blue}{0.750}  & 23.50   \\
                            & RefineDNet~\cite{zhao2021refinednet} & TIP’2021  & 20.48    & 0.838    & 17.06    &                      & 20.87     & 0.895    & 10.28    &                      & 15.50   & 0.697  & \color{red}{21.90}  \\
                            & YOLY~\cite{li2021you}          & IJCV’2021 &15.50     &0.844     &22.21     &                      & 15.96     & 0.869    & 13.44     &                      &15.31    &0.647   &21.57   \\
                            & $D^{4}$~\cite{yang2022self}          & CVPR’2022 & 25.42    & 0.914    & 11.52    &                      & 25.87     & 0.947    & 5.21     &                      & \color{blue}15.70   & 0.737  & 23.01  \\
                            & $D^{4}+$~\cite{yang2023robust}        & IJCV’2023  & \color{blue}25.78    & \color{blue}{0.918}    & \color{blue}{11.26}    &                      & \color{blue}{26.29}     & \color{blue}{0.950}    & \color{blue}{5.15}     &                      & 15.60   & 0.732  & 22.89  \\
                            & \textbf{Ours}        &           & \textbf{\color{red}{26.23}}    & \textbf{\color{red}0.932}    & \textbf{\color{red}{9.20}}     &                      & \textbf{\color{red}{26.52}}     & \textbf{\color{red}{0.951}}    & \textbf{\color{red}{4.71}}     &                      & \textbf{\color{red}{16.23}}   & \textbf{\color{red}{0.752}}  & \textbf{\color{blue}{22.81}}  \\ \hline
\end{tabular}

\label{table:table1}
}
\end{table*}
To begin with, we convert the input image into a grayscale image denoted as $I_{gray}$. Subsequently, we apply Sobel's operator to calculate the horizontal and vertical gradients of the grayscale image, represented as $\nabla_{x}$ and $\nabla_{y}$, respectively.
\begin{flalign}\label{eq1}
&\boldsymbol \nabla_x(\boldsymbol I_{gray})=\boldsymbol I_{gray}*\boldsymbol K_x,&
\end{flalign}
\begin{flalign}\label{eq1}
&\boldsymbol \nabla_y(\boldsymbol I_{gray})=\boldsymbol I_{gray}*\boldsymbol K_y,&
\end{flalign}
\begin{flalign}\label{eq1}
&\boldsymbol H_{_{Sobel}}=\begin{bmatrix}-1&0&1\\-2&0&2\\-1&0&1\end{bmatrix},\quad \boldsymbol V_{_{Sobel}}=\begin{bmatrix}-1&-2&-1\\0&0&0\\1&2&1\end{bmatrix},&
\end{flalign}
where $K_x$ and $K_y$ denote the horizontal and vertical kernels of the Sobel operator, respectively.
We calculate the gradient magnitude $M$ based on the horizontal and vertical gradients. The final contour loss $L_{contour}$ consists of the average of the gradient magnitudes. It can be written as follows:
\begin{flalign}\label{eq1}
&\boldsymbol M=\sqrt{\nabla_x(\boldsymbol I_{gray})^2 + \nabla_y(\boldsymbol I_{gray})^2},&
\end{flalign}
\begin{flalign}\label{eq1}
&\boldsymbol L_{contour}=\frac1N\sum_{i=1}^NM_i,&
\end{flalign} 
where $ N $ denotes the total number of pixels in the image.

Based on the above considerations and analyses, the total loss function in our experiment can be expressed as:
\begin{flalign}\label{eq1}
&\boldsymbol L_{total}=\lambda_1 \boldsymbol L_{cyc} + \lambda_2 \boldsymbol L_{adv} + \lambda_3 \boldsymbol L_{contour}.&
\end{flalign} 
where $\lambda_i$ represents different loss weights. In our experiments, we set $\lambda_1$ = 1, $\lambda_2$ = 1 and $\lambda_3$ = 0.5 according to the experimental settings.
	\par{
		
	}
	\section{EXPERIMENTS}
	\label{EXPERIMENTS}
 \textbf{Implementation Details.} The overall architecture of the DCM-dehaze, and during training, we use the Adam optimizer~\cite{kingma2014adam} with $\beta_1$ = 0.9, $\beta_2$ = 0.999, and a batch size of 2. The model has a learning rate of 0.0001. In addition, we randomly crop the images into 256 × 256 patches for training. The entire network is executed on an NVIDIA 3090 GPU using the PyTorch framework.\par
\textbf{Datasets.} To train the model, we use the RESIDE~\cite{li2018benchmarking} datasets. RESIDE's Indoor Training Datasets (ITS) contain 13990 synthetic hazy images and 1399 clean images. We evaluate the proposed method on synthetic datasets and real-world datasets. The synthetic datasets containing 500 indoor and 500 outdoor images were used for the test datasets, along with I-haze~\cite{ancuti2018haze} datasets containing 35 images.\par
\textbf{Evaluation Metric.}~\autoref{table:table1} shows the quantitative results of different dehazing methods on the SOTS-indoor, SOTS-outdoor, and I-haze datasets. To evaluate the performance of our method, we chose these three metrics for quantitative comparison: PSNR, SSIM~\cite{wang2004image} and CIEDE2000~\cite{sharma2005ciede2000}. These metrics are commonly used to evaluate the effectiveness of dehazing networks.
\begin{table*}[width=\textwidth,ht!]
\caption{Ablation study on DCM-dehaze. " \ding{55} "(resp. " \checkmark~") means the corresponding module is unused(resp. used). The first and second performances are shown in {\color{red} red} and {\color{blue} blue}, respectively.}
\scalebox{0.90}{
\renewcommand\arraystretch{1.2}
\begin{tabular}{lllllllllllllll}
\hline
\multicolumn{4}{l}{Methods} & \multicolumn{3}{l}{SOTS-indoor} & \multicolumn{1}{c}{} & \multicolumn{3}{l}{SOTS-outdoor} & \multicolumn{1}{c}{} & \multicolumn{3}{l}{I-haze} \\ \cline{5-7} \cline{9-11} \cline{13-15} 
DDSCM  & DFRE  & ATT  & BCA  & PSNR$\uparrow$      & SSIM$\uparrow$     & CIEDE$\downarrow$    &                      & PSNR$\uparrow$      & SSIM$\uparrow$      & CIEDE$\downarrow$     &                      & PSNR$\uparrow$    & SSIM$\uparrow$    & CIEDE$\downarrow$   \\ \hline
      \ding{55} &\ding{55}       &\ding{55}      &\ding{55}      & 25.42     & 0.914    & 11.52    &                      & 25.87     & 0.947     & 5.21     &                      & 15.70   & 0.737   & 23.01  \\
     \checkmark &\ding{55}       &\ding{55}      &\ding{55}      & 25.66     & 0.919    & 11.35    &                      & 25.96     & \color{red}0.951     & 5.14     &                      & 15.93   & 0.748   & 22.45  \\
      \ding{55} &\checkmark       &\ding{55}      &\ding{55}      & 25.53     & 0.917    & 9.84     &                      & 25.97     & \color{blue}0.950     & 5.21     &                      & \color{blue}16.22   & \color{blue}0.751   & \color{blue}21.93  \\
     \ding{55} &\checkmark       &\ding{55}      &\checkmark      & 25.59     & \color{blue}0.926    & 10.38    &                      & 25.98     & \color{blue}0.950     & \color{blue}5.02     &                      & 15.98   & \color{red}0.752   & \color{red}21.69  \\
      \ding{55}&\checkmark       &\checkmark      &\ding{55}      & \color{blue}26.14     & 0.924    & \color{blue}9.83     &                      & \color{blue}26.06     & 0.947     & 5.20     &                      & 16.11   & 0.749   & 22.02  \\
    \checkmark  &\ding{55}       &\checkmark      &\checkmark      & 26.00     & 0.925    & 9.93     &                      & 25.94     & 0.947     & 5.34     &                      & 15.15   & 0.717   & 22.72  \\ 
    \checkmark  &\checkmark       &\checkmark      &\checkmark      & \color{red}26.23     & \color{red}0.932    & \color{red}9.20     &                      & \color{red}26.52     & \color{red}0.951     & \color{red}4.71     &                      & \color{red}16.23   & \color{red}0.752   & 22.81  \\ \hline
\end{tabular}
\label{table:table2}
}
\end{table*}
\begin{table}[]
\caption{Comparison to w/o DDSCM (BCA, DFRE) module on SOTS datasets. The \textbf{bold font} indicates the best performance.}
\scalebox{0.72}{
\renewcommand\arraystretch{1.2}
\begin{tabular}{llllllll}
\hline
Methods   & \multicolumn{3}{l}{SOTS-indoor} & \multicolumn{1}{c}{} & \multicolumn{3}{l}{SOTS-outdoor} \\ \cline{2-4} \cline{6-8} 
      & PSNR$\uparrow$      & SSIM$\uparrow$     & CIEDE$\downarrow$    &                      & PSNR$\uparrow$      & SSIM$\uparrow$      & CIEDE$\downarrow$    \\ \hline
w/o DDSCM & 25.81     & 0.923    & 11.63    &                      & 25.95     & 0.949     & 4.85     \\
w/o BCA   & 24.55     & 0.915    & 12.27    &                      & 24.34     & 0.933     & 6.83     \\
w/o DFRE  & 25.86     & 0.924    & 10.07     &                      & 25.94     & 0.947     & 5.38     \\
\textbf{Ours}      & \textbf{26.23}     & \textbf{0.932}    & \textbf{9.20}     &                      & \textbf{26.52}     & \textbf{0.951}     & \textbf{4.71}     \\ \hline
\end{tabular}
}
\label{table:table3}
\end{table}
\subsection{Comparisons on real-world hazy images} To further evaluate the haze removal performance of natural scenes, we conducted experiments on SOTS-indoor and SOTS-outdoor datasets. For AODNet~\cite{li2017aod}, EPDN~\cite{qu2019enhanced}, FFANet~\cite{qin2020ffa}, HardGAN~\cite{deng2020hardgan}, PSD~\cite{chen2021psd}, CycleGAN~\cite{yan2023ade}, RefineDNet~\cite{zhao2021refinednet}, YOLY~\cite{li2021you}, $D^{4}$~\cite{yang2022self}, $D^{4}+$~\cite{yang2023robust}, we used their published model that was pre-trained on authentic haze images. ~\autoref{fig:fig6} shows the visualized results of the SOTS-outdoor datasets. \par\par
\textbf{Quantitative evaluation for image dehazing} As shown in~\autoref{table:table1}, Quantitative evaluation consisted of ten methods, with the initial five being supervised and the latter five unsupervised. We used the three evaluation metrics, PSNR, SSIM, and CIEDE, as shown in~\autoref{table:table1}, compared to the original model, the PSNR index on the SOTS-indoor, SOTS-outdoor and I-haze datasets increased by 0.81 dB, 0.65 dB and 0.53 dB, respectively. The CIEDE index fell 2.32 dB, 0.50 dB, and -0.20 dB, respectively.\par
\textbf{Qualitative evaluation for image dehazing.} We further show the dehazing results of five outdoor haze images in~\autoref{fig:fig6} for qualitative comparison. From these visual results, we can easily observe that FFANet~\cite{qin2020ffa} and RefineDNet~\cite{zhao2021refinednet} make the dehaze effect too light and too heavy, respectively. For $D^4$~\cite{yang2022self}, we find that it cannot completely remove the deep residual haze. Although quite good dehaze can be achieved in some cases, our DCM-dehaze is one of the best, maintaining the original brightness while removing as much of the input haze as possible.
\subsection{Ablation Study}
To demonstrate the effectiveness of our proposed DCM-dehaze, an ablation study is performed next, and the main experiments analyzed include Bidirectional Contour Analysis, Attention mechanism, Dual Depthwise Separable Convolutional Module, and Dense Flow Residual Enhancer.

\begin{figure}[htbp]
\centering 
\includegraphics[width=\linewidth]{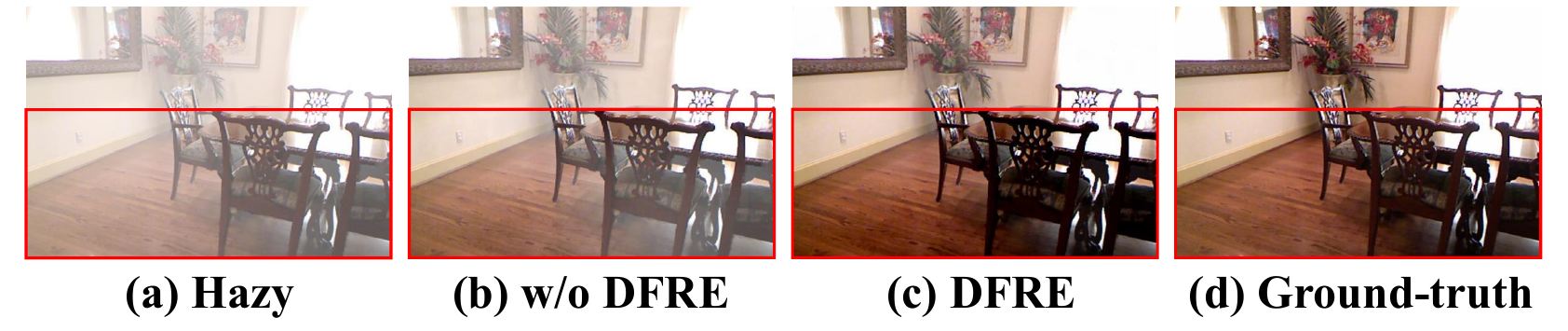}
\caption{A visual comparison of a real image from the SOTS-indoor datasets. The DFRE module has a relatively positive impact on the haze removal effect. }
\label{fig:fig7}
\end{figure}

\textbf{Effect of DFRE Module.} As shown in~\autoref{table:table2}, for the I-haze datasets, the PSNR of the DFRE module is significantly improved by 0.62 dB from basic to "basic + DFRE", so DFRE removes some redundant features compared to the original feature information because it has a higher performance gain. Furthermore, as shown in~\autoref{table:table3}, compared to the absence of the DFRE module, the PSNR indices in the indoor and outdoor SOTS data sets increased by 0.37 dB and 0.58 dB, respectively, while the CIEDE indices decreased by 0.87 dB and 0.67 dB, respectively. ~\autoref{fig:fig7} shows an intuitive visual contrast, which ultimately helps to restore clarity and detail to the dehazing image.
\par
\textbf{Effect of DDSCM Module.} As shown in \autoref{table:table2}, there is still a significant improvement in the SOTS-outdoor dataset. It indicates that the DDSCM module enhances the detailed feature response of the image by transmitting information to deeper levels of the network.\par
\textbf{Effect of BCA Module.} From~\autoref{table:table3}, we can see that any ablation to our network causes a noticeable drop in performance. Compared with the network without bidirectional contours analysis, our complete network DCM-dehaze dehaze is more thorough. The PSNR and SSIM indexes were also increased by 1.68 dB and 2.18 dB in the SOTS-indoor and SOTS-outdoor datasets. Therefore, the BCA module facilitates the enhancement of model performance. 
	\section{Conclusion}
	\label{Conclusion}
	\par{
		This paper proposes a novel dehazing method (DCM-dehaze). The framework consists of dehazing and contour constraint branches. Specifically, we optimize the edge information of the image using a bidirectional contour analysis method to enhance the clarity of the details and thus improve fidelity. Meanwhile, the Dense Flow Residual Enhancer removes known redundant features when recovering a clear image. Also, We utilize a bidirectional separable convolution module to alleviate the problem of detail feature distortion due to domain shifts. The recovered image is made more realistic. Experimental results show that the method effectively removes hazed images in synthetic and natural worlds.
	}
 \section*{Acknowledgement}
This work was supported in part by the Chongqing University of Technology high-quality development Action Plan for graduate education gzlcx20242046, the Basic and Applied Basic Research Foundation of Guangdong Province under Project 2021A1515110298, and in part by the Science and Technology Program of Nansha under Project 2021ZD003.

\begin{thebibliography}{10}

\bibitem{an2022semi}
Shunmin An, Xixia Huang, Le~Wang, Linling Wang, and Zhangjing Zheng.
\newblock Semi-supervised image dehazing network.
\newblock {\em The Visual Computer}, 38(6):2041--2055, 2022.

\bibitem{ancuti2018haze}
Cosmin Ancuti, Codruta~O Ancuti, Radu Timofte, and Christophe De~Vleeschouwer.
\newblock I-haze: a dehazing benchmark with real hazy and haze-free indoor images.
\newblock In {\em Advanced Concepts for Intelligent Vision Systems: 19th International Conference, ACIVS 2018, Poitiers, France, September 24--27, 2018, Proceedings 19}, pages 620--631. Springer, 2018.

\bibitem{ref3}
Dana Berman, Shai Avidan, et~al.
\newblock Non-local image dehazing.
\newblock In {\em Proceedings of the IEEE conference on computer vision and pattern recognition}, pages 1674--1682, 2016.

\bibitem{chen2021psd}
Zeyuan Chen, Yangchao Wang, Yang Yang, and Dong Liu.
\newblock Psd: Principled synthetic-to-real dehazing guided by physical priors.
\newblock In {\em Proceedings of the IEEE/CVF conference on computer vision and pattern recognition}, pages 7180--7189, 2021.

\bibitem{cong2024semi}
Xiaofeng Cong, Jie Gui, Jing Zhang, Junming Hou, and Hao Shen.
\newblock A semi-supervised nighttime dehazing baseline with spatial-frequency aware and realistic brightness constraint.
\newblock In {\em Proceedings of the IEEE/CVF Conference on Computer Vision and Pattern Recognition}, pages 2631--2640, 2024.

\bibitem{deng2020hardgan}
Qili Deng, Ziling Huang, Chung-Chi Tsai, and Chia-Wen Lin.
\newblock Hardgan: A haze-aware representation distillation gan for single image dehazing.
\newblock In {\em European conference on computer vision}, pages 722--738. Springer, 2020.

\bibitem{guo2022image}
Chun-Le Guo, Qixin Yan, Saeed Anwar, Runmin Cong, Wenqi Ren, and Chongyi Li.
\newblock Image dehazing transformer with transmission-aware 3d position embedding.
\newblock In {\em Proceedings of the IEEE/CVF conference on computer vision and pattern recognition}, pages 5812--5820, 2022.

\bibitem{guo2023haze}
Fan Guo, Jianan Yang, Zhuoqun Liu, and Jin Tang.
\newblock Haze removal for single image: A comprehensive review.
\newblock {\em Neurocomputing}, 2023.

\bibitem{ref1}
Kaiming He, Jian Sun, and Xiaoou Tang.
\newblock Single image haze removal using dark channel prior.
\newblock {\em IEEE transactions on pattern analysis and machine intelligence}, 33(12):2341--2353, 2010.

\bibitem{hodges2019single}
Cameron Hodges, Mohammed Bennamoun, and Hossein Rahmani.
\newblock Single image dehazing using deep neural networks.
\newblock {\em Pattern Recognition Letters}, 128:70--77, 2019.

\bibitem{kingma2014adam}
Diederik~P Kingma and Jimmy Ba.
\newblock Adam: A method for stochastic optimization.
\newblock {\em arXiv preprint arXiv:1412.6980}, 2014.

\bibitem{li2017aod}
Boyi Li, Xiulian Peng, Zhangyang Wang, Jizheng Xu, and Dan Feng.
\newblock Aod-net: All-in-one dehazing network.
\newblock In {\em Proceedings of the IEEE international conference on computer vision}, pages 4770--4778, 2017.

\bibitem{li2018benchmarking}
Boyi Li, Wenqi Ren, Dengpan Fu, Dacheng Tao, Dan Feng, Wenjun Zeng, and Zhangyang Wang.
\newblock Benchmarking single-image dehazing and beyond.
\newblock {\em IEEE Transactions on Image Processing}, 28(1):492--505, 2018.

\bibitem{li2021you}
Boyun Li, Yuanbiao Gou, Shuhang Gu, Jerry~Zitao Liu, Joey~Tianyi Zhou, and Xi~Peng.
\newblock You only look yourself: Unsupervised and untrained single image dehazing neural network.
\newblock {\em International Journal of Computer Vision}, 129:1754--1767, 2021.

\bibitem{li2019generative}
Ce~Li, Xinyu Zhao, Zhaoxiang Zhang, and Shaoyi Du.
\newblock Generative adversarial dehaze mapping nets.
\newblock {\em Pattern Recognition Letters}, 119:238--244, 2019.

\bibitem{li2020netnet}
Yazhao Li, Yanwei Pang, Jianbing Shen, Jiale Cao, and Ling Shao.
\newblock Netnet: Neighbor erasing and transferring network for better single shot object detection.
\newblock In {\em Proceedings of the IEEE/CVF conference on computer vision and pattern recognition}, pages 13349--13358, 2020.

\bibitem{liu2023nighthazeformer}
Yun Liu, Zhongsheng Yan, Sixiang Chen, Tian Ye, Wenqi Ren, and Erkang Chen.
\newblock Nighthazeformer: Single nighttime haze removal using prior query transformer.
\newblock In {\em Proceedings of the 31st ACM International Conference on Multimedia}, pages 4119--4128, 2023.

\bibitem{mahaur2023small}
Bharat Mahaur and KK~Mishra.
\newblock Small-object detection based on yolov5 in autonomous driving systems.
\newblock {\em Pattern Recognition Letters}, 168:115--122, 2023.

\bibitem{qin2020ffa}
Xu~Qin, Zhilin Wang, Yuanchao Bai, Xiaodong Xie, and Huizhu Jia.
\newblock Ffa-net: Feature fusion attention network for single image dehazing.
\newblock In {\em Proceedings of the AAAI conference on artificial intelligence}, volume~34, pages 11908--11915, 2020.

\bibitem{qu2019enhanced}
Yanyun Qu, Yizi Chen, Jingying Huang, and Yuan Xie.
\newblock Enhanced pix2pix dehazing network.
\newblock In {\em Proceedings of the IEEE/CVF conference on computer vision and pattern recognition}, pages 8160--8168, 2019.

\bibitem{ren2016single}
Wenqi Ren, Si~Liu, Hua Zhang, Jinshan Pan, Xiaochun Cao, and Ming-Hsuan Yang.
\newblock Single image dehazing via multi-scale convolutional neural networks.
\newblock In {\em Computer Vision--ECCV 2016: 14th European Conference, Amsterdam, The Netherlands, October 11-14, 2016, Proceedings, Part II 14}, pages 154--169. Springer, 2016.

\bibitem{sharma2005ciede2000}
Gaurav Sharma, Wencheng Wu, and Edul~N Dalal.
\newblock The ciede2000 color-difference formula: Implementation notes, supplementary test data, and mathematical observations.
\newblock {\em Color Research \& Application: Endorsed by Inter-Society Color Council, The Colour Group (Great Britain), Canadian Society for Color, Color Science Association of Japan, Dutch Society for the Study of Color, The Swedish Colour Centre Foundation, Colour Society of Australia, Centre Fran{\c{c}}ais de la Couleur}, 30(1):21--30, 2005.

\bibitem{shi2023cnn}
Runwu Shi, Shichun Yang, Yuyi Chen, Rui Wang, Mengyue Zhang, Jiayi Lu, and Yaoguang Cao.
\newblock Cnn-transformer for visual-tactile fusion applied in road recognition of autonomous vehicles.
\newblock {\em Pattern Recognition Letters}, 166:200--208, 2023.

\bibitem{song2023vision}
Yuda Song, Zhuqing He, Hui Qian, and Xin Du.
\newblock Vision transformers for single image dehazing.
\newblock {\em IEEE Transactions on Image Processing}, 32:1927--1941, 2023.

\bibitem{sun2023scale}
Hang Sun, Yan Zhang, Peng Chen, Zhiping Dan, Shuifa Sun, Jun Wan, and Weisheng Li.
\newblock Scale-free heterogeneous cyclegan for defogging from a single image for autonomous driving in fog.
\newblock {\em Neural Computing and Applications}, pages 1--15, 2023.

\bibitem{tarel2009fast}
Jean-Philippe Tarel and Nicolas Hautiere.
\newblock Fast visibility restoration from a single color or gray level image.
\newblock In {\em 2009 IEEE 12th international conference on computer vision}, pages 2201--2208. IEEE, 2009.

\bibitem{wang2024ucl}
Yongzhen Wang, Xuefeng Yan, Fu~Lee Wang, Haoran Xie, Wenhan Yang, Xiao-Ping Zhang, Jing Qin, and Mingqiang Wei.
\newblock Ucl-dehaze: Towards real-world image dehazing via unsupervised contrastive learning.
\newblock {\em IEEE Transactions on Image Processing}, 2024.

\bibitem{wang2004image}
Zhou Wang, Alan~C Bovik, Hamid~R Sheikh, and Eero~P Simoncelli.
\newblock Image quality assessment: from error visibility to structural similarity.
\newblock {\em IEEE transactions on image processing}, 13(4):600--612, 2004.

\bibitem{wu2021contrastive}
Haiyan Wu, Yanyun Qu, Shaohui Lin, Jian Zhou, Ruizhi Qiao, Zhizhong Zhang, Yuan Xie, and Lizhuang Ma.
\newblock Contrastive learning for compact single image dehazing.
\newblock In {\em Proceedings of the IEEE/CVF conference on computer vision and pattern recognition}, pages 10551--10560, 2021.

\bibitem{xue2024low}
Minglong Xue, Jinhong He, Yanyi He, Zhipu Liu, Wenhai Wang, and Mingliang Zhou.
\newblock Low-light image enhancement via clip-fourier guided wavelet diffusion.
\newblock {\em arXiv preprint arXiv:2401.03788}, 2024.

\bibitem{yan2023ade}
Bingnan Yan, Zhaozhao Yang, Huizhu Sun, and Conghui Wang.
\newblock Ade-cyclegan: A detail enhanced image dehazing cyclegan network.
\newblock {\em Sensors}, 23(6):3294, 2023.

\bibitem{yang2023adversarial}
Jenny Yang, Andrew~AS Soltan, David~W Eyre, Yang Yang, and David~A Clifton.
\newblock An adversarial training framework for mitigating algorithmic biases in clinical machine learning.
\newblock {\em NPJ digital medicine}, 6(1):55, 2023.

\bibitem{yang2023robust}
Yang Yang, Chaoyue Wang, Xiaojie Guo, and Dacheng Tao.
\newblock Robust unpaired image dehazing via density and depth decomposition.
\newblock {\em International Journal of Computer Vision}, pages 1--21, 2023.

\bibitem{yang2022self}
Yang Yang, Chaoyue Wang, Risheng Liu, Lin Zhang, Xiaojie Guo, and Dacheng Tao.
\newblock Self-augmented unpaired image dehazing via density and depth decomposition.
\newblock In {\em Proceedings of the IEEE/CVF conference on computer vision and pattern recognition}, pages 2037--2046, 2022.

\bibitem{ye2024low}
Jiongkai Ye, Yong Wu, and Dongliang Peng.
\newblock Low-quality image object detection based on reinforcement learning adaptive enhancement.
\newblock {\em Pattern Recognition Letters}, 182:67--75, 2024.

\bibitem{zhang2018densely}
He~Zhang and Vishal~M Patel.
\newblock Densely connected pyramid dehazing network.
\newblock In {\em Proceedings of the IEEE conference on computer vision and pattern recognition}, pages 3194--3203, 2018.

\bibitem{zhang2024depth}
Yafei Zhang, Shen Zhou, and Huafeng Li.
\newblock Depth information assisted collaborative mutual promotion network for single image dehazing.
\newblock In {\em Proceedings of the IEEE/CVF Conference on Computer Vision and Pattern Recognition}, pages 2846--2855, 2024.

\bibitem{zhao2021refinednet}
Shiyu Zhao, Lin Zhang, Ying Shen, and Yicong Zhou.
\newblock Refinednet: A weakly supervised refinement framework for single image dehazing.
\newblock {\em IEEE Transactions on Image Processing}, 30:3391--3404, 2021.

\bibitem{zhu2017unpaired}
Jun-Yan Zhu, Taesung Park, Phillip Isola, and Alexei~A Efros.
\newblock Unpaired image-to-image translation using cycle-consistent adversarial networks.
\newblock In {\em Proceedings of the IEEE international conference on computer vision}, pages 2223--2232, 2017.

\bibitem{ref2}
Qingsong Zhu, Jiaming Mai, and Ling Shao.
\newblock A fast single image haze removal algorithm using color attenuation prior.
\newblock {\em IEEE transactions on image processing}, 24(11):3522--3533, 2015.

\end{thebibliography}

\end{sloppypar}
\end{CJK}
\end{document}